\documentclass[11pt,a4paper]{article}
\usepackage[hyperref]{acl2021}
\usepackage{times}
\usepackage{latexsym}

\usepackage{amsmath}
\usepackage{amssymb}
\usepackage{caption}
\usepackage{subcaption}
\usepackage{enumerate}
\usepackage{multirow}
\usepackage{graphicx} 
\usepackage{url}
\usepackage{booktabs}
\usepackage{color}

\usepackage[utf8]{inputenc}
 
\usepackage{cleveref}
\crefname{section}{§}{§§}
\Crefname{section}{§}{§§}

\usepackage{microtype}

\aclfinalcopy 
\def\aclpaperid{722} 

\title{Logic-Driven Context Extension and Data Augmentation \\ for Logical Reasoning of Text}

\author{Siyuan Wang\textsuperscript{\rm 1}\thanks{Work is done during internship at Microsoft Research Asia.}, Wanjun Zhong\textsuperscript{\rm 2}, Duyu Tang\textsuperscript{\rm 3}, Zhongyu Wei\textsuperscript{\rm 1}, \\ \bf Zhihao Fan\textsuperscript{\rm 1}, Daxin Jiang\textsuperscript{\rm 3}, Ming Zhou\textsuperscript{\rm 4}, Nan Duan\textsuperscript{\rm 3}  \\
  \textsuperscript{\rm 1}School of Data Science, Fudan University, China \\
  \textsuperscript{\rm 2}Sun Yat-Sen University, China \textsuperscript{\rm 3}Microsoft, China
  \textsuperscript{\rm 4}Sinovation Ventures, China \\
  \texttt{\{wangsy18,zywei,fanzh18\}@fudan.edu.cn} \\
  \texttt{zhongwj25@mail2.sysu.edu.cn}; \texttt{zhouming@chuangxin.com} \\
  \texttt{\{dutang,djiang,nanduan\}@microsoft.com} \\
  }

\date{}

\begin{document}
\maketitle
\begin{abstract}
Logical reasoning of text requires understanding critical logical information in the text and performing inference over them. Large-scale pre-trained models for logical reasoning mainly focus on word-level semantics of text while struggling to capture symbolic logic. In this paper, we propose to understand logical symbols and expressions in the text to arrive at the answer. Based on such logical information, we not only put forward a context extension framework but also propose a data augmentation algorithm. The former extends the context to cover implicit logical expressions following logical equivalence laws. The latter augments literally similar but logically different instances to better capture logical information, especially logical negative and conditional relationships.
We conduct experiments on ReClor dataset. The results show that our method achieves the state-of-the-art performance, and both logic-driven context extension framework and data augmentation algorithm can help improve the accuracy. And our multi-model ensemble system is the first to surpass human performance on both EASY set and HARD set of ReClor. \footnote{Codes are publicly available at \url{https://github.com/WangsyGit/LReasoner}.} 
\end{abstract}

\section{Introduction}

Recent years have witnessed a growing interest in logical reasoning of text, which aims to follow a logic-level analysis over a given text to arrive at the correct answer \cite{mccarthy1989artificial, nilsson1991logic}.
It is a challenging task since it requires the ability to extract critical information from text and perform logical inference over them \cite{williams2018broad, habernal2018argument, liu2020logiqa}.




\begin{figure}[!th]
\centering
\includegraphics[width=1.0\columnwidth]{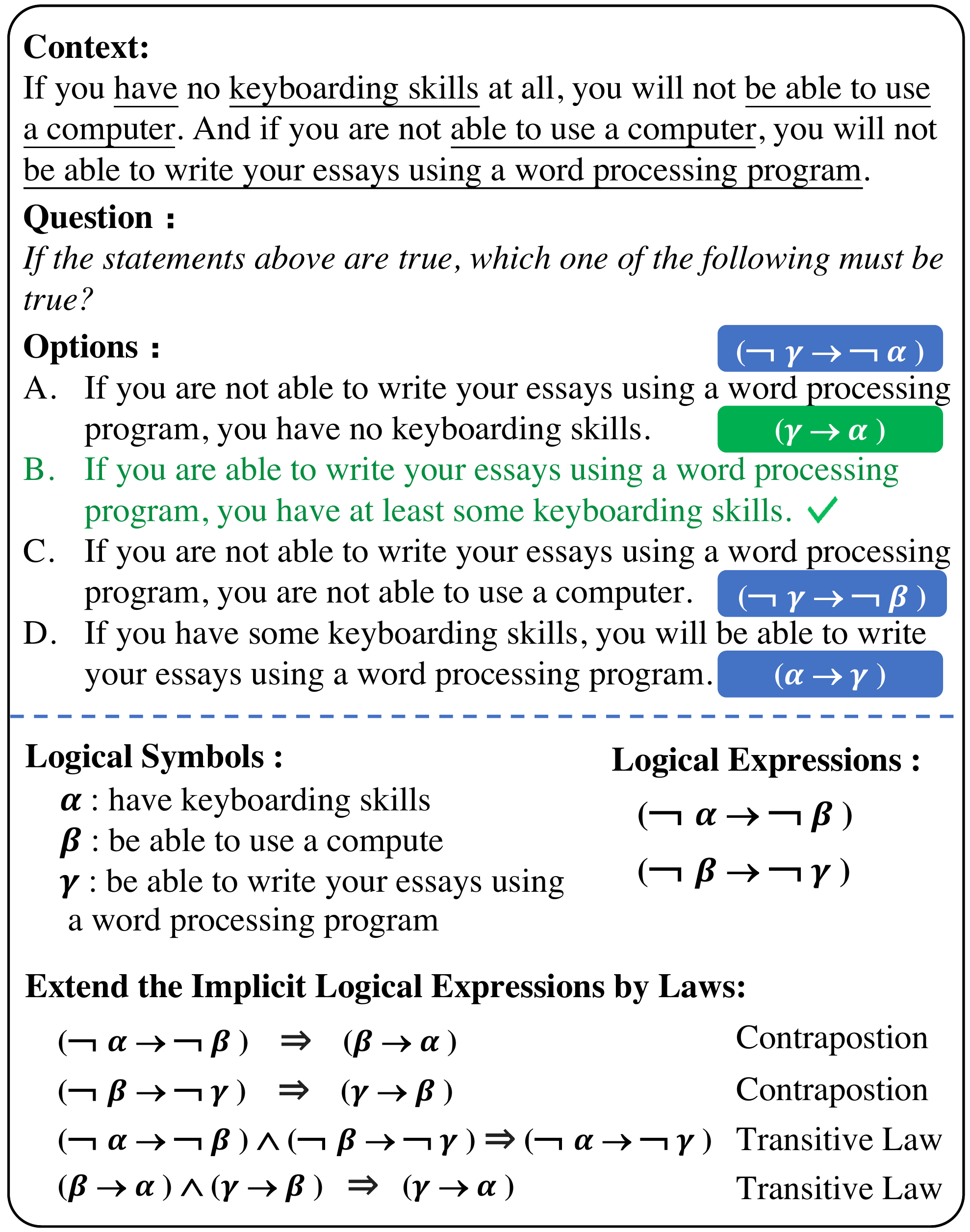}
\caption{\label{figure1} A logical reasoning example from ReClor dataset ~\cite{yu2020reclor} which includes a context, a question and four options and the correct answer is marked by \textcolor[RGB]{74,163,87}{\checkmark}. To find the answer, it need to extract \emph{logical symbols}, identify \emph{logical expressions} and perform logical inference to extend the \emph{implicit logical expressions}. The underlined phrases represent logical symbols. The colored rectangles are corresponding logical expressions of options.}
\end{figure}
An example of logical reasoning problem is shown in Figure~\ref{figure1}. It takes a context, a related question and four options as the input, and needs to analyze which option follows logically from the context.
In order to solve such a problem, the reasoning system first needs to extract the critical constituents from the context as logical symbols like $\alpha, \beta, \gamma$ and identify the existing logical expressions that are composed of logical symbols, like $(\neg \alpha \rightarrow \neg \beta)$.
Then according to logical equivalence laws, it performs inference to extend the logical expressions that are not explicitly mentioned in the context and compare them with the logical expressions of each option to find
the most plausible answer.




Although large-scale pre-trained language models \cite{devlin2019bert, liu2019roberta, yang2019xlnet} have achieved promising results on logical reasoning of text, they mainly leverage word-level semantics but hardly capture symbolic logic.
In responding to such a problem, we propose to identify logical symbols and expressions explicitly mentioned in the context which serve as the elementary components of logical inference process.
Based upon such logical information, we propose both a context extension framework and a data augmentation algorithm. 
The whole system is built on top of the pre-trained model considering its promising performance.
Our logic-driven context extension framework performs inference following logical equivalence laws to extend the logical expressions in the context to cover the implicit ones.
For better utilization of extended logical expressions at symbolic space into neural model, we verbalize them into natural language and feed as an extended context into a pre-trained model to match the options and find the answer.
Our logic-driven data augmentation algorithm employs modification operations over logical expressions to produce challenging instances with literally similar but logically different contexts. It trains our model to distinguish different contexts to better capture logical information, especially negative and conditional relationships in logical expressions.

We take both RoBERTa \cite{liu2019roberta} and ALBERT \cite{lan2019albert} as our backbone pre-trained models. The experiments are conducted on the benchmark dataset ReClor \cite{yu2020reclor} and our system achieves the state-of-the-art performance and reaches the top of the leaderboard. 
The further results also show the effectiveness of both logic-driven context extension framework and data augmentation algorithm.

\section{Task and Background}
\subsection{Task Definition}
\label{task_definition}
We study the problem of logical reasoning of text on a multiple-choice question answering task. The task is described as following: given a context $c$, a question $q$, and four associated options $\{o_1,o_2,o_3,o_4\}$, we aim to select the most appropriate option as the answer $o_a$. 

We adopt a challenging dataset ReClor \cite{yu2020reclor} which is collected from the logical reasoning questions of standardized examinations including Law School Admission Test (LSAT)\footnote{\url{https://www:lsac:org/lsat/taking-lsat/testformat/logical-reasoning}} and Graduate Management Admission Test (GMAT)\footnote{\url{https://www.mba.com/exams/gmat/about-the-gmat-exam/gmat-exam-structure/verbal}}. 
Although many questions need to be answered through complicated logical reasoning, there are still some biased instances which can be solved even without knowing contexts and questions.
It therefore splits the unbiased instances from the test data as the HARD set to fully evaluate the logical reasoning ability. And the other biased ones are taken as the EASY set. A leaderboard\footnote{\url{https://eval.ai/web/challenges/challenge-page/503/leaderboard/1347}} is also hosting for public evaluation on ReClor.

\begin{figure*}[!th]
\centering
\includegraphics[width=2.08\columnwidth]{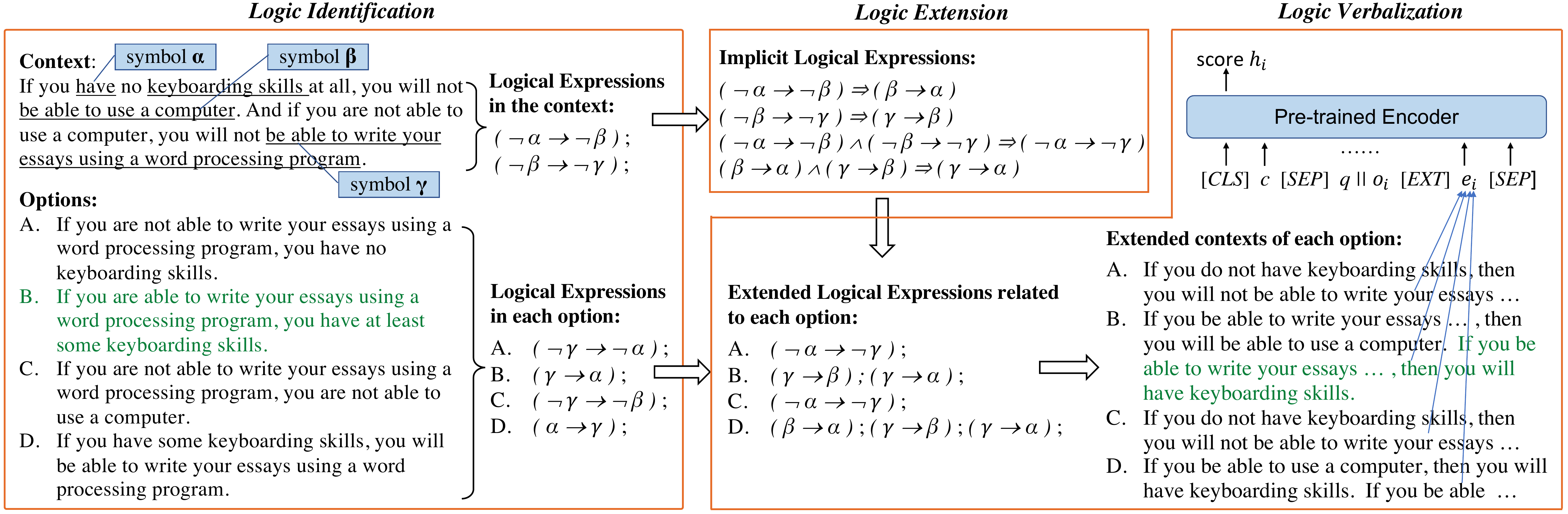}
\caption{\label{figure_framework} The overall architecture of our proposed logic-driven context extension framework. $c$, $q$, $o_i$ and $e_i$ are the context, question, $i$-th option and the extended context for $i$-th option, respectively.
The texts in \textcolor[RGB]{74,163,87}{green} mean that the option $B$ is matched against its extended context which has the highest score.}
\end{figure*}
\subsection{Base Model}
\label{section_base_model}
In this paper, we follow the leading methods on the leaderboard to take pre-trained models as our base model.
Pre-trained models for multiple-choice question answering concatenate the context, the question and each option as the input and encode the sequence for calculating its score. Given four options, four concatenated sequences are constructed to calculate four scores, and the one with the highest score is chosen as the answer. 
Specifically, the concatenated sequence is formulated as $[CLS] \ c \ [SEP] \ q \ || \ o \ [SEP]$, where $c$ is the context and $q \ || \ o$ is the concatenation of the question and each option.
The representations of the special token $[CLS]$ in the four sequences are fed into a linear layer with a softmax function to get the probability distribution of options as $P(\{o_1,o_2,o_3,o_4\}|c,q)$. 
The cross entropy loss is calculated as:
\begin{align}
    \mathcal{L}_A = - \sum \mathop{\log}P(o_a|c,q) 
    \label{loss_1}
\end{align}
where $o_a$ is the correct option.

Since previous work \cite{yu2020reclor} has shown that RoBERTa \cite{liu2019roberta} achieves superior results on the leaderboard, and ALBERT \cite{lan2019albert} is proved to be even more powerful than RoBERTa on many tasks, we utilize both RoBERTa and ALBERT in the experiments.
\paragraph{RoBERTa} is a robustly optimized BERT \cite{devlin2019bert} with more training data, which uses a more dynamic sentence masking method and removes the next sentence prediction loss.
\paragraph{ALBERT} is a lite BERT with two parameter-reduction techniques including factorized embedding parameterization and cross-layer parameter sharing. And it replaces the next sentence prediction loss with the inter-sentence coherence loss. 

However, such pre-trained models for logical reasoning of text directly encode the triplet of context, question and options, which only leverage word-level semantics but struggle to capture symbolic logic in the text.
Therefore, we propose a framework on top of a pre-trained model to understand logical symbols and expressions in the text and perform logical inference over them to predict the answer.

\section{Logic-Driven Context Extension}
In this section, we present a logic-driven context extension framework for logical reasoning of text, and the overall architecture is illustrated in Figure~\ref{figure_framework}. 
The framework can be divided into three steps as follow. It first identifies the logical symbols and expressions explicitly mentioned in the context and options
(\cref{sec:identification_step}). Then it performs logical inference over them to extend the logical expressions implicit in the context (\cref{sec:extension_step}). Finally, it verbalizes the extended logical expressions related to each option as an extended context and utilizes it in the pre-trained model to match the answer (\cref{sec:translation_step}).

\subsection{Logic Identification}
\label{sec:identification_step}
In order to perform logical reasoning, we first need to identify the elementary components for reasoning as logical expressions. 
We identify the existing logical expressions for each sentence in the context and each option. 
To show the format of the logical expression, we introduce some notations:
\begin{enumerate}[(1)]
    \setlength{\itemsep}{1pt}
    \item $\{\alpha, \beta, \gamma,...\}$: the logical symbols, which are the basic constituents in the context to constitute the logical expressions, such as the ``have keyboarding skills" in Figure~\ref{figure_framework}.
    \item $\{\neg, \rightarrow \}$: the logical connectives set. $\neg$ means the negation operation upon a specific logical symbol and $\rightarrow$ acts as a conditional relationship between two logical symbols.
    \item $\{(\alpha \rightarrow \beta), ...\}$: the logical expressions which are composed of logical symbols and connectives. $(\alpha \rightarrow \beta)$ means that $\alpha$ is the condition of $\beta$.
\end{enumerate}

Because logical expressions are composed of logical symbols, we first employ constituency parser~\cite{joshi2018extending} to extract constituents including noun phrases and gerundial phrases as basic symbols.
Then the logical symbols in each sentence are combined by logical connectives to constitute the logical expressions as follow-up. If any negative word is related to a logical symbol $\alpha$, we add the negation connective $\neg$ before $\alpha$ as a new logical symbol $\neg \ \alpha$.
We define a set of negative words as \{``\emph{not}", ``\emph{n't}", ``\emph{unable}", ``\emph{no}", ``\emph{few}", ``\emph{little}", ``\emph{neither}", ``\emph{none of}"\}.
Then if there is a conditional relationship between two logical symbols $\alpha$ and $\beta$ in a sentence, we can construct the corresponding logical expression as $(\alpha \rightarrow \beta)$.
We simply recognize the conditional relationship between symbol $\alpha$ and $\beta$ as $(\alpha \rightarrow \beta)$ according to conditional indicators such as \emph{``if $\alpha$, then $\beta$", ``$\alpha$ in order for $\beta$", ``$\beta$ due to $\alpha$", ``$\neg \beta$ unless $\alpha$"}, etc.
And if an active voice occurs between $\alpha$ and $\beta$, we also have $(\alpha \rightarrow \beta)$. As illustrated in Figure~\ref{figure_framework}, given the context with two sentences, we can extract three logical symbols $\{\alpha, \beta, \gamma\}$ and identify two existing logical expressions as $(\neg \alpha \rightarrow \neg \beta)$ and $(\neg \beta \rightarrow \neg \gamma)$.


\subsection{Logic Extension}
\label{sec:extension_step}
In addition to the logical expressions explicitly mentioned in the context, there are still some other implicit ones which we need to logically infer and extend.
We combine the identified logical expressions existing in all sentences of the context as a logical expression set $\mathcal{S}$, and perform logical inference over them to further extend the implicit logical expressions according to logical equivalence laws. Here we follow two logical equivalence laws including \emph{contraposition} \cite{russel2013artificial} and \emph{transitive law} \cite{zhao1997static}:
\begin{align}
    \text{Contraposition}: & \nonumber \\
    (\alpha \rightarrow \beta) & \implies (\neg \beta \rightarrow \neg \alpha)  \\
    \text{Transitive Law}: & \nonumber \\
    (\alpha \rightarrow \beta) \ \land \ & (\beta \rightarrow \gamma) \implies (\alpha \rightarrow \gamma)  
\end{align}
Then the extended implicit logical expressions form an extension set of the current logical expression set $\mathcal{S}$ as $\mathcal{S}_E$.

As the example in Figure~\ref{figure_framework}, the set of existing logical expressions is $\mathcal{S}=\{(\neg \alpha \rightarrow \neg \beta), (\neg \beta \rightarrow \neg \gamma)\}$ and the logic extension set is $\mathcal{S}_E=\{(\beta \rightarrow \alpha), (\gamma \rightarrow \beta), (\neg \alpha \rightarrow \neg \gamma), (\gamma \rightarrow \alpha)\}$.

\subsection{Logic Verbalization}
\label{sec:translation_step}
After inferring the extended logical expressions set $\mathcal{S}_E$, we verbalize them into natural language for better utilization in the pre-trained model.
We first select the related expressions from $\mathcal{S}_E$ for each option. A logical expression is regarded as related to an option if it has the same logical symbols with the option judged by the text overlapping, and whether a negation connective exists also needs to be considered. 
For example, $(\neg \alpha \rightarrow \neg \gamma)$ in Figure~\ref{figure_framework} is related to option $C$ because they both contain $\neg \gamma$. Then we transform all logical expressions related to the option at symbolic space into natural language by filling them into a template and concatenate them into a sentence. We take such a sentence as an extended context for this option. The template is designed as shown in Table~\ref{table_template}.
\begin{table}[ht]
\begin{center}
\begin{tabular}{|p{1.3cm}|p{5.5cm}|}
\hline
logic & $(\neg \alpha \rightarrow \neg \gamma)$ \\
\hline
template & If do not $\alpha$, then will not $\gamma$. \\
\hline
extended context & If you do not have keyboarding skills, then you will not be able to write your essays using a word processing program.\\
\hline
\end{tabular}
\end{center}
\caption{\label{table_template} An example of verbalizing a logical expression into text.}
\end{table}

We feed extended contexts into the pre-trained model to match the options and predict the answer. 
We take an extended context as sentence $e$, and introduce a special token $[EXT]$ to represent context extension. Then we reformulate the input sequence as
$[CLS] \ c \ [SEP] \ q \ || \ o \ [EXT] \ e \ [SEP]$ for encoding and feed the $[CLS]$ representation into a classification layer to get each option's score and find the most appropriate answer.  

\section{Logic-Driven Data Augmentation}
In order to make the pre-trained model put more focus on logical information in the context, especially logical negative and conditional relationships, we further introduce a logic-driven data augmentation algorithm. Inspired by SimCLR \cite{chen2020simple}, it employs contrastive learning to augment challenging instances with literally similar but logically different contexts based on logical expressions. It then trains our model to predict the correct context supporting the answer.
We first introduce the background of SimCLR and then describe our logic-driven constrative learning. 

\paragraph{SimCLR}
As a paradigm of self-supervised representation learning by comparing different samples,
contrastive learning  \cite{wu2018unsupervised, le2020contrastive, he2020momentum} aims to make the representations of similar samples be mapped close together, while that of dissimilar samples be further away in the encoding space. The goal can be described as following.
\begin{align}
    s(f(x), f(x^+)) \gg s(f(x), f(x^-))
\end{align}
$x^+$ is a positive sample similar to the data point $x$ while $x^-$ is a negative sample dissimilar to $x$. $f(\cdot)$ is an encoder to learn a representation and the $s(\cdot)$ is a metric function that measures the similarity between two representations. 
Over this, SimCLR \cite{chen2020simple} bulids a classifier to distinguish positive from negative samples and learns to capture what makes two samples different. 

\paragraph{Logic-Driven Contrastive Learning} In our multiple-choice question answering setting, we alter the score function from measuring the similarity between two representations to calculating the score that the question can be solved by the correct answer under a given context:
\begin{align}
    s^{'}(c^+, q, o_a) \gg s^{'}(c^-, q, o_a)
\end{align}
where $(c^+, q, o_a)$ and $(c^-, q, o_a)$ are the positive and negative sample, $c^+$ and $c^-$ are the positive and negative context, respectively, and $s^{'}$ is the score function.
The contrastive loss can be formulated as a classification loss for predicting the most plausible context that supports the answer:
\begin{align}
    \mathcal{L}_C = -\sum \log\frac{\exp(s^{'}(+))}{\exp(s^{'}(+))+\exp(s^{'}(-))}
\end{align}
where $s^{'}(+)$ and $s^{'}(-)$ are short for $s^{'}(c^+, q, o_a)$ and $s^{'}(c^-, q, o_a)$ respectively. 

\begin{figure}[!th]
\centering
\includegraphics[width=0.9\columnwidth]{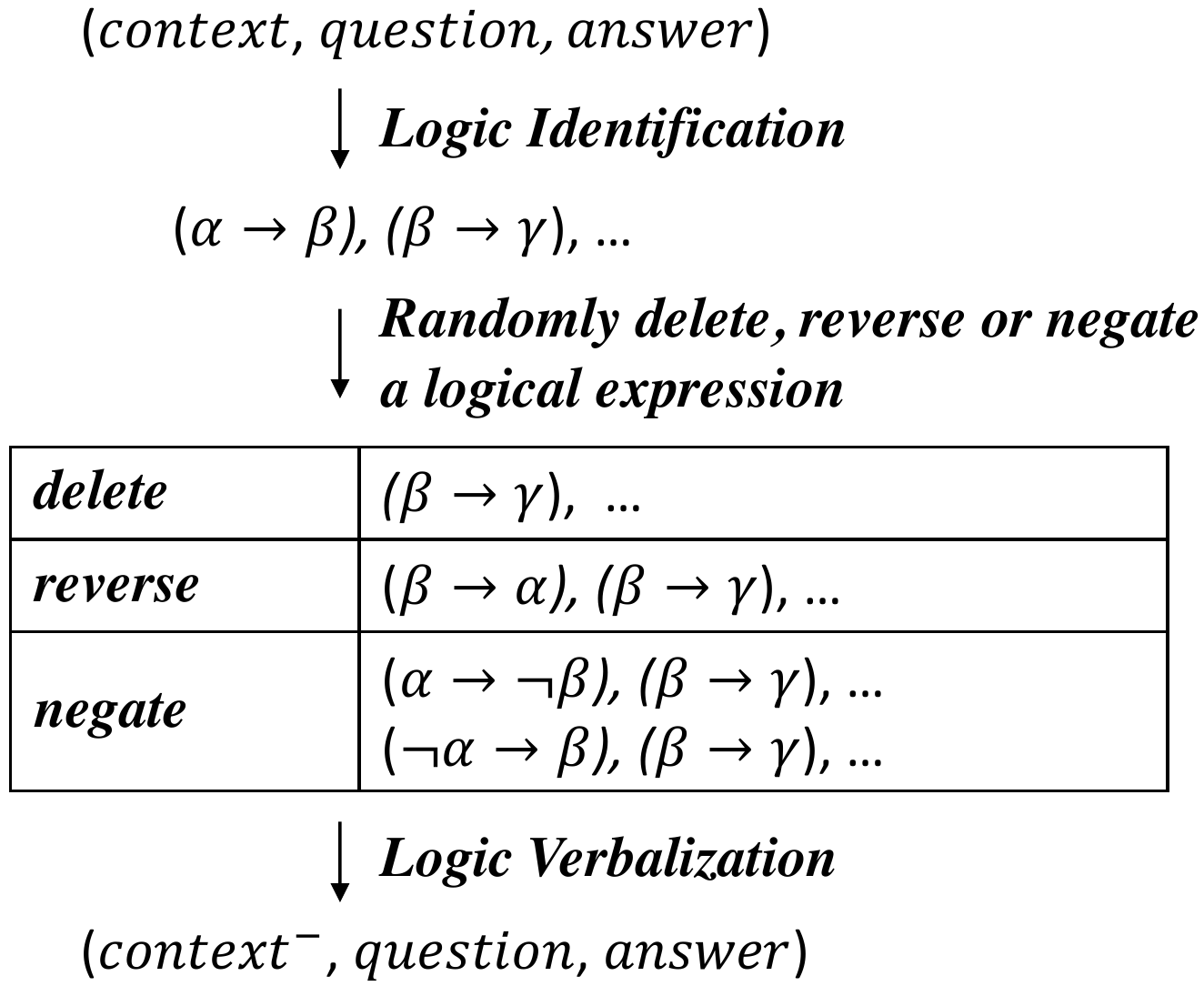}
\caption{\label{figure_contrastive} Procedure to construct a logical negative sample.}
\end{figure}

Aware of symbolic logical expressions, we can construct \emph{logical negative samples} including negative contexts that are literally similar but logical dissimilar to the positive one. 
We take the original context to construct the positive sample. Then we generate a negative sample by modifying the existing logical expressions in the context and verbalizing the modified logical expressions into a negative context as \cref{sec:translation_step}. During the modification operations, we randomly choose a logical expression and randomly \emph{delete, reverse or negate} such a expression. The \emph{delete, reverse or negate} operations are respectively to delete a logical expression in the context, reverse the conditional order of a logical expression and negate a logical symbol in a logical expression.
The constructing procedure of a logical negative sample is illustrated in Figure~\ref{figure_contrastive}. Then the model can be trained to better capture logical information, especially negative and conditional relationships in logical expressions.

In the logic-driven data augmentation algorithm, our framework is trained with a combined loss as $\mathcal{L}=\mathcal{L}_A+\mathcal{L}_C$. And the classification of positive and negative context for the correct answer is also implemented in the logic-driven context extension framework.

\section{Experiments}
\subsection{Experimental Setup}
Our experiments are conducted on ReClor~\cite{yu2020reclor}, a question answering dataset extracted from standardized exams including GMAT and LSAT which requires complicated logical reasoning. It consists of $6,138$ questions and each question is collected with a context and four answer options, in which only one is correct. It is divided into training, validation and test sets with $4,638$, $500$ and $1,000$ data points, and test set is split into EASY set and HARD set with $440$ and $560$ data points. For automatic evaluation, we adopt the accuracy as the metric to compare performance of models over test set, EASY set and HARD set.
 
We take both RoBERTa-large \cite{liu2019roberta} and ALBERT-xxlarge-v2 \cite{lan2019albert} as our backbone models and implement them using Huggingface \cite{wolf2019huggingface}. We use a batch size of 8 and fine-tune on ReClor for 10 epochs. The AdamW \cite{loshchilov2018fixing} with $\beta1$ = 0.9 and $\beta2$ = 0.98 is taken as the optimizer and the learning rate is set to 1e-5. We use a linear learning rate scheduler with $10\%$ warmup proportion.
We select at most two extended logical expressions related to each option to construct the extended context. 

\subsection{Overall Performance}
We compare our systems with several baseline models and human performance, which are described as follows.
\paragraph{Baseline Models} The compared baseline pre-trained models for multiple-choice question answering include GPT \cite{radford2018improving}, GPT-2 \cite{radford2019language}, BERT \cite{devlin2019bert}, XLNet \cite{yang2019xlnet}, RoBERTa \cite{liu2019roberta} and ALBERT \cite{lan2019albert}. In this paper, BERT, XLNet, RoBERTa and ALBERT respectively refer to BERT-large, XLNet-large, RoBERTa-large and ALBERT-xxlarge-v2.
\paragraph{Our Systems} \emph{$\text{LReasoner}_{\text{RoBERTa}}$} is our proposed \underline{\bf l}ogic-driven \underline{\bf reasoner} built on top of RoBERTa, which utilizes both logic-driven context extension framework and data augmentation algorithm. \emph{$\text{LReasoner}_{\text{ALBERT}}$} is also our proposed \underline{\bf l}ogic-driven \underline{\bf reasoner} but taking ALBERT as the backbone model. 
\paragraph{Human Performance} The average score of ten graduate students over randomly chosen 100 samples from test set is taken as human performance.
\begin{table}[!th]
    \setlength\tabcolsep{5pt}
    \newcommand{\tabincell}[2]{\begin{tabular}{@{}#1@{}}#2\end{tabular}}
	\centering
	\begin{tabular}{c|ccc}
    \toprule
    \bf Model & \bf Test & \bf EASY & \bf HARD \\
	\midrule 
    $\text{GPT}^*$ & 45.4 & 73.0 & 23.8 \\
    $\text{GPT-2}^*$ & 47.2 & 73.0 & 27.0 \\
    $\text{BERT}^*$ & 49.8 & 72.0 & 32.3 \\
    $\text{XLNet}^*$ & 56.0 & 75.7 & 40.5 \\
	\midrule 
    $\text{RoBERTa}^*$ & 55.6 & 75.5 & 40.0 \\
    $\text{LReasoner}_{\text{RoBERTa}}$ & 62.4 & \bf 81.4 & 47.5 \\
	\midrule 
    ALBERT & 66.5 & 76.6 & 58.6 \\
    $\text{LReasoner}_{\text{ALBERT}}$ & \bf 70.7 & 81.1 & 62.5 \\
	\midrule  
	$\text{Human Performance}^*$ & 63.0 & 57.1 & \bf 67.2 \\
	\bottomrule
 	\end{tabular}
	\caption{Experimental results of different models and human performance. The results in \textbf{bold} are the best performance of each column and $*$ indicates that the results come from \citet{yu2020reclor}.}
	\label{table_result_1}
\end{table}

The evaluation results are shown in Table~\ref{table_result_1}. We have several findings:
\begin{enumerate}[-]
    \setlength{\itemsep}{1pt}
    \setlength{\parskip}{1pt}

    \item Our models outperform all baseline models by a considerable margin. \emph{$\text{LReasoner}_{\text{ALBERT}}$} even surpasses the human performance in \emph{Test} and \emph{EASY} sets.
    This indicates the effectiveness of our method for predicting more accurate answer.
    
    \item \emph{$\text{LReasoner}_{\text{RoBERTa}}$} and \emph{$\text{LReasoner}_{\text{ALBERT}}$} both perform better than the corresponding baseline models \emph{RoBERTa} and \emph{ALBERT}. It demonstrates that our proposed method is robust to be effective for logical reasoning on top of different pre-trained models, even the base model is already of great power, such as \emph{ALBERT}.
    
    \item Our models generate large improvement on both \emph{HARD} set and \emph{EASY} set by comparing our models
    with \emph{RoBERTa} and \emph{ALBERT}, respectively. This not only follows our intuition that our system is designed for logical reasoning problems, but also shows that it is capable of solving easier problems. 
    
\end{enumerate}

\subsection{Ablation Study}
To dive into the effectiveness of different components in our logic-driven reasoner, we conduct ablation study which takes \emph{RoBERTa} as our backbone model on both validation and test sets.
\begin{table}[!th]
    \newcommand{\tabincell}[2]{\begin{tabular}{@{}#1@{}}#2\end{tabular}}
	\centering
	\begin{tabular}{l|cccc}
	\toprule
    \bf Model & \bf Val & \bf Test & \bf EASY & \bf HARD \\
	\midrule
    $\text{RoBERTa}$ & 62.6 & 55.6 & 75.5 & 40.0 \\
    \ \ \ + $\text{CE}$ & 65.2 & 58.3 & 78.6 & 42.3 \\
    \ \ \ + $\text{DA}$ & 65.8 & 61.0 & 80.9 & 45.4 \\
    \ \ \ + $\text{CE}$ + $\text{DA}$ & 66.2 & 62.4 & 81.4 & 47.5 \\
	\bottomrule
 	\end{tabular}
	\caption{Ablation study of our system built on top of \emph{RoBERTa}. \emph{$\text{CE}$} and \emph{$\text{DA}$} are respectively our logic-driven \underline{\bf c}ontext \underline{\bf e}xtension framework and \underline{\bf d}ata \underline{\bf a}ugmentation algorithm. \emph{$\text{RoBERTa}$+$\text{CE}$+$\text{DA}$} is our proposed \emph{$\text{LReasoner}_{\text{RoBERTa}}$}.}
	\label{table_ablation}
\end{table}

As shown in Table~\ref{table_ablation}, \emph{$\text{RoBERTa}$+$\text{CE}$} and \emph{$\text{RoBERTa}$+$\text{DA}$} both outperform the baseline model \emph{RoBERTa} and perform worse than our final system \emph{$\text{RoBERTa}$+$\text{CE}$+$\text{DA}$}. It indicates that both logic-driven context extension framework and logic-driven data augmentation algorithm are beneficial for question answering involving logical reasoning.

\subsection{Further Analysis on Negative Sample Construction Strategies}
To further analyze the effectiveness of our logical negative samples in logic-driven contrastive learning, we compare several different negative sample construction strategies in contrastive learning on top of \emph{RoBERTa}.
\begin{table}[!th]
    \setlength\tabcolsep{4pt}
    \newcommand{\tabincell}[2]{\begin{tabular}{@{}#1@{}}#2\end{tabular}}
	\centering
	\begin{tabular}{l|ccc}
	\toprule
    \bf Model & \bf Test & \bf EASY & \bf HARD \\
	\midrule
    $\text{RoBERTa (w/o CLR)}$ & 55.6 & 75.5 & 40.0 \\
    $\text{RoBERTa (w/ CLR-RS)}$ & 58.2 & 79.3 & 41.6 \\
    $\text{RoBERTa (w/ CLR-RD)}$ & 58.9 & 78.9 & 43.2 \\
    $\text{RoBERTa (w/ CLR-L)}$ & 61.0 & 80.9 & 45.4 \\
	\bottomrule
 	\end{tabular}
	\caption{Comparison of different negative sample construction approaches on top of \emph{RoBERTa}.
	\emph{CLR} represents contrastive learning. \emph{RS} means \underline{\bf r}andomly \underline{\bf s}electing a context from in-batch data while \emph{RD} means \underline{\bf r}andomly \underline{\bf d}eleting a sentence from the original context as the negative sample. \emph{L} denotes our \underline{\bf l}ogical negative samples in logic-driven contrastive learning.}
	\label{table_logical_negative}
\end{table}

\begin{figure*}[!th]
\centering
\includegraphics[width=2.07\columnwidth]{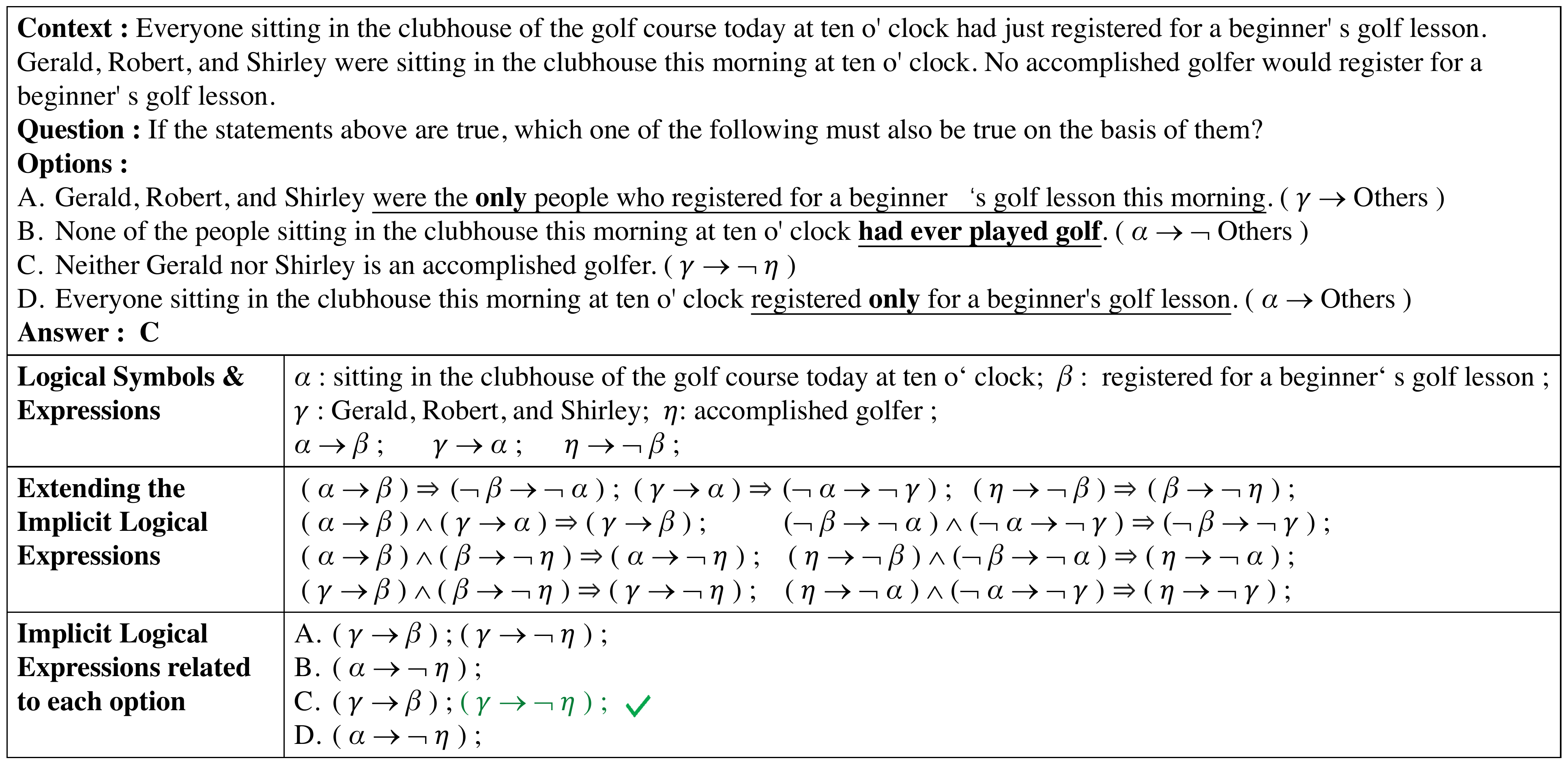}
\caption{\label{figure_case} A case of the reasoning process of our logic-driven reasoner \emph{$\text{LReasoner}_{\text{ALBERT}}$}. Phrases underlined represent other symbols (called \emph{Others}) different from the logical symbols in context and tokens in \textbf{bold} make them different. The option marked by \textcolor[RGB]{74,163,87}{\checkmark} is our predicted answer.}
\end{figure*}

From the results in Table~\ref{table_logical_negative}, we can see that all models with contrastive learning outperform the model without it, which demonstrates that contrastive learning can help to better predict the answer. Our logic-driven contrastive learning \emph{RoBERTa(w/ CLR-L)} performs best. It reveals that logical negative samples are more effective than negative samples constructed by other methods which make the model to better capture the logical negative and conditional relationships in the context for logical reasoning.

\subsection{Case Study}
A case study is presented in Figure~\ref{figure_case} to show the reasoning process of our system. At first the logical symbols are correctly extracted from the context and the logical expressions are identified based on them considering logical negative and conditional relationships. Then we extend the logical expressions by inferring implicit ones in the context. For each option we recognize its logical expression and find the related extended expressions. We verbalize them into natural language to feed into pre-trained model as an extended context to compute a matching score. Finally we take option C which exactly matches an extended implicit logical expression as the most plausible answer. 

\subsection{Error Analysis}
\begin{table}[!h]
    \setlength\tabcolsep{3pt}
    \newcommand{\tabincell}[2]{\begin{tabular}{@{}#1@{}}#2\end{tabular}}
	\centering
	\begin{tabular}{lcc}
	\toprule
	\bf Reasoning Type & \bf Base & \bf Ours \\
	\midrule  				
	Necessary Assumptions (11.0\%)& 73.7 & 76.3 ($\uparrow$)\\
	Sufficient Assumptions (3.6\%)& 70.0 & 70.0 ($-$)\\
	Strengthen (9.0\%)& 69.1 & 70.2 ($\uparrow$)\\
	Weaken (10.6\%)& 64.6 & 59.3 ($\downarrow$)\\
	Evaluation (1.6\%)& 69.2 & 69.2 ($-$)\\
	Implication (6.2\%)& 43.8 & 54.3 ($\uparrow$)\\
	Conclusion/Main Point (3.1\%)& 80.6 & 77.8 ($\downarrow$)\\
	Most Strongly Supported (6.7\%)& 58.9 & 71.4 ($\uparrow$)\\
	Explain or Resolve (8.0\%)& 60.7 & 67.9 ($\uparrow$) \\
	Principle (5.7\%) & 72.3 & 76.9 ($\uparrow$)\\ 
	Dispute (2.5\%) & 63.3 & 80.0 ($\uparrow$)\\
	Technique (3.8\%) & 75.0 & 80.6 ($\uparrow$)\\
    Role (3.7\%) & 78.1 & 68.8 ($\downarrow$)\\
    Identify a Flaw (11.3\%) & 65.0 & 71.8 ($\uparrow$)\\
    Match Flaws (4.9\%) & 61.3 & 61.3 ($-$)\\
    Match the Structure (2.7\%) & 56.7 & 86.7 ($\uparrow$)\\
    Others (5.5\%) & 68.5 & 72.6 ($\uparrow$)\\
	\bottomrule
 	\end{tabular}
	\caption{Detailed results on different logical reasoning types. Numbers in parentheses are percentages of different reasoning types. \textbf{Base} is the \emph{ALBERT} model while \textbf{Ours} means our \emph{$\text{LReasoner}_{\text{ALBERT}}$} system.
	$\uparrow$, $\downarrow$ and $-$ respectively mean that our performance is better, worse than and equal to the baseline model.}
	\label{table_type}
\end{table}


Although our system achieves the best performance, there still exists some instances that can not be solved. 
ReClor dataset integrates various logical reasoning skills which can be categorized into 17 types. We can investigate the detailed performance with respect to different logical reasoning types of our system \emph{$\text{LReasoner}_{\text{ALBERT}}$} and the baseline model \emph{ALBERT}, to analyse which type of questions tend to be more challenging.

As shown in Table~\ref{table_type}, we can see that all models perform relative poorly on certain logical reasoning types, such as \texttt{Most Strongly Supported} and \texttt{Implication}. The first type is less straight-forward and requires deeper logical reasoning considering degree. Specifically, \texttt{Most Strongly Supported} aims to find the choice that is \textbf{most} strongly supported by a stimulus. For \texttt{Implication}, all candidate options prefer to be more similar to each other or the context which are more difficult to be classified, like the examples in Figure~\ref{figure1} and~\ref{figure_case}. But our system is still able to make progress on them which prove the effectiveness of our system for logical reasoning.

Our model achieves great improvement in most of the logical reasoning types compared to the baseline model but performs comparably or even worse in following question types: \texttt{Sufficient Assumption}, \texttt{Evaluation}, \texttt{Conclusion/Main Point}, \texttt{Role}, \texttt{Match flaws} and \texttt{Weaken}. The first four types only take up a small proportion of the whole data, which results in no improvement over them. 
\texttt{Match flaws} needs to understand more abstract concept \textbf{flaw} and find an option exhibiting the same flaw as the context.
\texttt{Weaken} aims to find the opposite statement that \textbf{weaken} the argument.
Our system to first extract logical expressions and then imply the implicit logical expressions is not suitable for matching flaw and identifying weaken statement.
How to deal with different logical reasoning types is our further research direction.

\section{Related Work}
In recent years, an increasing number of tasks and datasets have been introduced targeting on logical reasoning of text in NLP area. Natural Language Inference (NLI) \cite{dagan2005pascal} is a typical task requiring logical reasoning, which aims to determine 
whether a hypothesis can reasonably be entailed from a premise. SNLI \cite{bowman2015large}, MultiNLI \cite{williams2018broad}, QNLI \cite{wang2018glue} and SciTail \cite{khot2018scitail} 
are widely used benchmarks in evaluating the performance of NLI. However, these datasets handle the task at sentence-level and the required logical reasoning ability is simple. \citet{liu2020natural} propose ConTRoL dataset investigating NLI for long texts and examining more complex contextual reasoning types. Argument reasoning comprehension \cite{habernal2018argument} is a task closer to NLI which also concerns passage-level logical reasoning. Given a premise and a claim, it aims to identify the correct implicit warrant from two opposing options. But it is limited to only one logical reasoning type, i.e., warrant identification. 

Another task involving logical reasoning of text is question answering. Several multiple-choice question answering datasets, which need to select an answer from candidate options given a context and a question, have been proposed for promoting the development of logical reasoning. These datasets are all sourced from public standardized exams so that they partly mitigate the noise brought by crowd-sourcing or automatically-generation \cite{lai2017race}. LogiQA \cite{liu2020logiqa} is collected from National Civil Servants Examination of China which covers 5 types of logical reasoning. \citet{yu2020reclor} propose ReClor dataset from the GMAT and LSAT tests. It examines more complicated logical reasoning integrating 17 question types. In this paper, we conduct experiments on top of ReClor for investigating diverse logical reasoning skills. 

Large-scale pre-trained language models \cite{devlin2019bert, liu2019roberta, yang2019xlnet, lan2019albert} have been widely adopted for logical reasoning of text, which directly encode the given texts to predict the output. Without understanding symbolic logic, they can still achieve a promising performance. We therefore take a pre-trained model as our backbone model and identify logical expressions for our proposed logic-driven system .

\section{Conclusion and Future Work}
In this paper, we focus on the task of logical reasoning of text and propose to understand the logical symbols and expressions in the context to find the answer. Taking them as elementary components of logical inference, we first propose a logic-driven context extension framework to extend the logical expressions to cover the implicit ones and verbalize them as an extended context to match the answer. We also introduce a logic-driven data augmentation algorithm which employs contrastive learning to augment literally similar but logically different instances to help our model better capture logical information, especially logical negative and conditional relationships. Experimental results on ReClor dataset confirm the effectiveness of our logic-driven reasoner with both logic-driven context extension framework and data augmentation algorithm.

In the future, we will explore how to take different logical reasoning types into consideration to deal with the question types that still can not be well solved. We also would like to design a model to directly encode the symbolic logic rather than utilize them by verbalizing into natural language. Besides, we plan to evaluate our model on more datasets. As our codes are publicly available, anyone interested 
could also have a try.

\bibliographystyle{acl_natbib}
\bibliography{anthology,acl2021}


\end{document}